\documentclass[conference]{IEEEtran}
\IEEEoverridecommandlockouts
\usepackage{cite}
\usepackage{amsmath,amssymb,amsfonts}
\usepackage{algorithmic}
\usepackage{graphicx}
\usepackage{textcomp}
\usepackage{xcolor}
\usepackage{listings}
\usepackage{url}
\usepackage[export]{adjustbox}
\usepackage{svg}
\usepackage{url}
\usepackage{hyperref}

\def\BibTeX{{\rm B\kern-.05em{\sc i\kern-.025em b}\kern-.08em
    T\kern-.1667em\lower.7ex\hbox{E}\kern-.125emX}}
\begin{document}

\title{LLM Reasoner and Automated Planner:\\ A new NPC approach
}

\author{\IEEEauthorblockN{1\textsuperscript{st} Israel Puerta-Merino}
\IEEEauthorblockA{\textit{
IIIA - Artificial Intelligence Research Institute
} \\
\textit{CSIC - Spanish Scientific Research Council
}\\
Bellaterra, Cataluña, España \\
israelpm01@hotmail.com}
\and
\IEEEauthorblockN{2\textsuperscript{nd} Jordi Sabater-Mir}
\IEEEauthorblockA{\textit{
IIIA - Instituto de Investigación en Inteligencia Artificial
} \\
\textit{CSIC - Consejo Superior de Investigaciones Científicas}\\
Bellaterra, Cataluña, España \\
jsabater@iiia.csic.es}
}

\maketitle

\begin{abstract}
In domains requiring intelligent agents to emulate plausible human-like behaviour, such as formative simulations, traditional techniques like behaviour trees encounter significant challenges. Large Language Models (LLMs), despite not always yielding optimal solutions, usually offer plausible and human-like responses to a given problem. In this paper, we exploit this capability and propose a novel architecture that integrates an LLM for decision-making with a classical automated planner that can generate sound plans for that decision. The combination aims to equip an agent with the ability to make decisions in various situations, even if they were not anticipated during the design phase.
\end{abstract}


\section{Introduction}
\label{sec:introduction}

Intelligent agents are required to emulate plausible human-like behaviour in multiple domains, such as serious games or formative simulations. Within this context, Non-Playable Characters (NPCs) usually face the challenge of making context-based decisions that must appear plausible and coherent for the human observer. Classical approaches to this problem, such as behaviour trees, encounter limitations. These methods require developers to anticipate and manually specify the actions and decision-making flow for every potential scenario, a task that is tedious, prone to errors and often hindered by the complexity of the simulated world. In response to these limitations, it would be desirable to have an architecture capable of making contextually appropriate decisions without exhaustive scenario specification.

On the other side, Large Language Models (LLMs) with recent papers such as Generative Agents \cite{generativeagents}, are recently demonstrating an interesting potential in decision-making problems resolution, being capable to emulate a plausible reasoning process. Despite this models are not always yielding reliable optimal solutions, they usually provide plausible and human-like responses to given problems.

In certain contexts, such as the one presented in the next section (\ref{sec:use_case}. Use Case), optimal behavior might not be necessary. In these domains, where we want NPCs to make decisions that are not necessarily optimal but coherent, it is worthwhile to explore the potential of an LLM as a decision-making reasoner, emulating plausible human-like behavior. This approach could lead to the development of a powerful reasoning system that emphasizes coherent behavior over high performance, making it suitable for scenarios where optimal processing is unnecessary but where a flexible model capable of delivering plausible and human-like responses is desired.

On the other hand, LLMs are computationally expensive and sometimes inconsistent, as they can generate hallucinations (i.e., incoherent or infeasible answers); additionally, they generate the answers in natural language, which can create difficulties in producing actionable plans.

\begin{figure}
    \centering
    \includegraphics[width=0.9\linewidth]{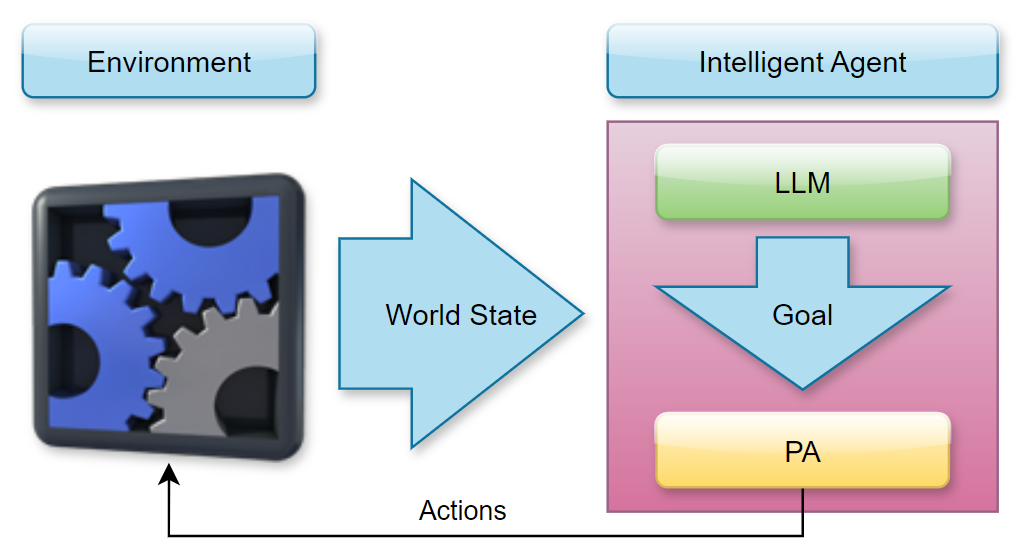}
    \caption{General scheme of the system behaviour. The intelligent agent receives the information about the world state and uses an LLM to generate the goal to follow; then uses PA techniques to, with the provided world state and the generated goal, find a list of actions necessary to achieve it.}
    \label{fig:systemGeneral}
\end{figure}

Therefore, to exploit capabilities of LLMs while minimizing their drawbacks, we propose a novel intelligent agent architecture that integrates an LLM for decision-making with a classical Automated Planning (AP) algorithm to generate sound plans that achieve the decisions. As shown in Fig. \ref{fig:systemGeneral}, the intelligent agent uses an LLM to decide which goal to follow based on the world state context. Then, it uses an Automated Planning (AP) algorithm to create a plan to achieve that goal - a list of specific actions that the environment interface can understand and sequentially execute. This combination aims to empower an autonomous agent with the flexibility to adapt to any situation, even those unforeseen by the developer, while maintaining plausible and human-like behavior.




\subsection{Use Case}
\label{sec:use_case}

\begin{figure}
    \centering
    \includegraphics[width=0.5\linewidth, frame=1pt]{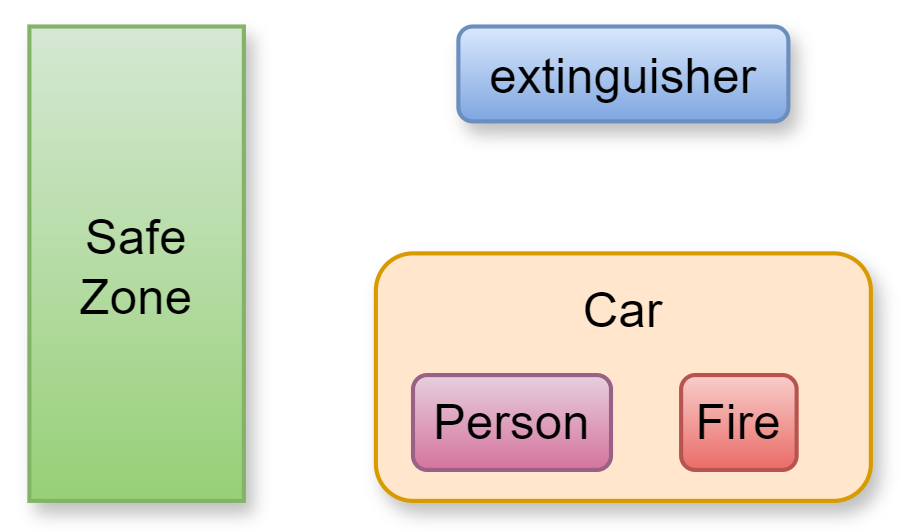}
    \caption{General scheme of \textit{The FireFighter Problem} Scenario.}
    \label{fig:firefighetScheme}
\end{figure}

This work is conducted under the \textit{RHYMAS: Real-time Hybrid Multiscale Agent-based Simulations for Emergency Training} (PID2020-113594RB-100) \cite{Sabater-Mir2022} project, a project in collaboration with the \textit{``Escola de Bombers, Protecció Civil i Agents Rurals de Catalunya,''} aimed at developing a framework for emergency simulations based on multiagent systems. Therefore, to illustrate how the developed system operates, we present a specific scenario called "The FireFighter Problem" (Fig. \ref{fig:firefighetScheme}). This scenario consists of three main components: a burning car with a person inside, an extinguisher nearby, and a safe zone. The person will be called \textit{Peter}, and the available actions for the agent in this domain are: moving from one place to another, taking or carrying something or someone, dropping the object or person it was carrying (if applicable), and extinguishing a fire with a fire extinguisher (only if carrying one).

In order to explore how the developed agent behaves with different roles (Firefighter, Paramedic, Common Person, etc.) and various personality traits, we have implemented this scenario that emulates a simple yet real dangerous situation. The implemented agent should be able to read the context, deduce the possible goals it can pursue, reasonably choose one of them, and make a plan to achieve it.

\subsection{System Description}
\label{sec:sys_desc}

\begin{figure*}
    \centering
    \includegraphics[width=0.85\linewidth]{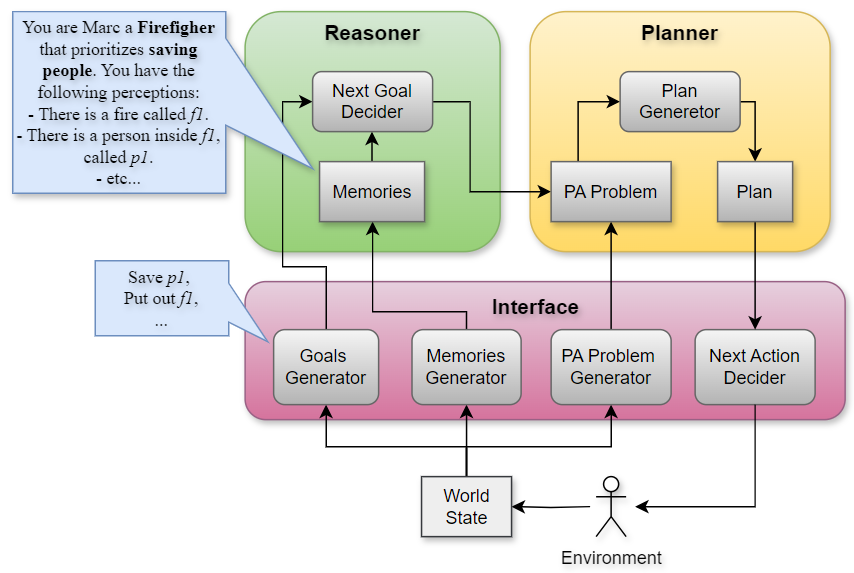}
    \caption{General architecture of the intelligent agent designed and implemented.}
    \label{fig:generalArchitecture}
\end{figure*}

The intelligent system developed employs an LLM to make the environment-based decision of which goal to follow. Subsequently, it utilizes the AP algorithm to devise an executable plan, a sequential list of specific actions to achieve the selected goal. As depicted in Figure \ref{fig:generalArchitecture}, the general architecture of this Intelligent Agent is based on three main modules: the Reasoner module, which utilizes the LLM to generate the goal; the Planner module, which employs AP techniques to generate the plan; and the Interface with the environment. Each module is briefly explained below:
\begin{itemize}

    \item \textbf{Reasoner}. This module features a main data structure, called \textit{memories}, representing the environmental context. These memories comprise the list of the pertinent perceptions the agent has encountered, presented in natural language format for the LLM to comprehend.

    The module receives a list of potential goals to pursue and, leveraging the environment context provided by the memories, determines which one to pursue.

    To enhance the model's capacity to address the reasoning process, we introduce an additional field in the Reasoner's memories, called the \textit{personality traits} field. This field serves as an initial phrase, also expressed in natural language, where aspects such as the agent's role in the environment or its priorities are specified.

    \item \textbf{Planner}. This module features a world-state representation called AP Problem, which employs a data structure compatible with the AP algorithm. 
    
    It receives the selected goal and generates a plan, a sequence of actions to achieve it in the current context (defined on the AP Problem).
    
    \item \textbf{Interface}. This module receives a new instance of the environment state and utilizes it to generate all the information required by the other modules. It is responsible for processing all the world-state information to generate (1) the possible goals that the agent could achieve in that specific situation,\footnote{The method to generate the possible goals depends on the specific environment and its implementation. In our \textit{Use Case}, we use an independent module for goal generation. This module utilizes a range of general goal classes, which receive the environment state and generate a list of achievable goals accordingly.} (2) the new perceptions to add to the Reasoner's memories, and (3) the AP Problem representation of the received world state. Once the plan is generated, the interface sequentially gives the actions to the environment to be executed.
\end{itemize}

The execution of the system consist on an iterative and reactive process where, on each iteration, the system exhibits the following behaviour:

\begin{enumerate}
    \item The iteration commences when the Interface module receives a new instance of the world-state. It utilizes the world-state information to generate all possible goals.
    \item The Interface module generates natural language perceptions of the world-state and compares them with the Reasoner`s memories, adding all the new ones. Subsequently, it constructs the AP Problem representation of the new world-state.
    \item If any new perception has been added to the memories, the Reasoner module re-evaluates the goal to achieve using the updated memories list and the goals generated by the interface.
    \item If the generated AP Problem differs from the previous one or the selected goal has changed, the Planner module generates a new plan.
    \item If the plan has changed or the agent has completed the previously selected action, the interface takes the action at the top of the plan, removes it from the list and sends it to the environment for execution.
\end{enumerate}

\subsection{Document Structure}

Finally, here is a brief summary of what can be found in the following pages:

\begin{itemize}
    \item \textbf{Chapter \ref{sec:art}. \textit{State Of The Art}}. We present a review of the fields of knowledge related to this work: NPC decision-making strategies, LLM research and AP resources. 

    \item \textbf{Chapter \ref{sec:design}.  \textit{Agent Architecture}}. We detail the design decisions made regarding the developed architecture and explain how each module works.
       
    \item \textbf{Chapter \ref{sec:tech}. \textit{Technologies Used}}. We present all the different technologies utilized in this project, providing explanations of their context and functionality.
    
    \item \textbf{Chapter \ref{sec:implementation}. \textit{Implementation and Testing}}. We document the programming process of the entire designed system, encompassing the integration of LLM and AP software, the implementation of the intelligent agent within a specific framework (Rhymas), the development of a specific environment (\textit{The FireFighter Problem}), and lastly, the testing of the agent's behavior in various contexts.

    \item \textbf{Chapter \ref{sec:discussion}. \textit{Discussions}}. We provide an exhaustive discussion about the performance of the agent, its capabilities and limitations, as well as the successes and failures that the LLMs still seem to have in this area of knowledge. Additionally, we discuss possible upgrades and propose future work based on this study.
    
    \item \textbf{Chapter \ref{sec:conclusion}. \textit{Conclusions}}. We conclude with a succinct summary of the work performed, highlighting key insights, findings, and significant conclusions drawn from our research endeavor.
\end{itemize}

\section{State of the Art}
\label{sec:art}

\subsection{NPC Behaviour Techniques}

The modeling of NPCs' behavior is a widely explored field with several contemporary approaches. In this article, we present a classification based on the one made by Andrey Simonov \cite{simonov2019applying}. 

The most primitive approach involves implementing predefined, usually cyclical behaviors. The result is a non-flexible and preplanned (i.e., nonexistent) decision-making process. This behavior might be convenient in some scenarios but is not useful for implementing autonomous or reactive agents.

This approach was soon improved with finite-state machines, which offer a series of predefined behaviors (states) but allow the agent's internal state (and, therefore, its behavior) to change depending on the environment context or trigger events. This provides more variable behavior and reactive capacity. A great example of this is an NPC that patrols an area but follows the player when discovered. However, while this works well for simple behaviors, the more states and triggers involved, the more implementation time is required. Moreover, their implementation is very closely tied to the environment's characteristics, making it a technique with low extensibility to different contexts. \cite{state_machine}

Behavior trees are hierarchical structures where each node represents a simple decision. When a node is activated, it makes its corresponding decision based on the environment context or internal states and activates the lower nodes related to the decision taken. Each node can execute a specific action when activated, but only the leaf nodes have behavior associated with them. This architecture represents a significant improvement over finite-state machines, allowing the implementation of the same behaviors in a simpler way. Additionally, it facilitates the creation of more complex behaviors due to the ease of constructing large tree structures. These structures are also more extensible, allowing for the creation of preset trees or blueprints that can be easily implemented in multiple environments. However, behavior trees still follow an action-reaction scheme, requiring the developer to explicitly define the entire agent's decision-making process. This makes it nearly impossible to implement a fully believable agent that considers all possible scenarios. Additionally, the large size of current trees makes them difficult to navigate and debug, leading to a higher likelihood of unexpected errors, especially in specific situations that might not have been contemplated during the design process. \cite{miyake2015current}

On the other hand, machine learning models like deep neural networks have gained strong prominence in recent decades, achieving great performance in playing various types of games. Additionally, Reinforcement Learning (RL) techniques have made the creation of machine learning agents plausible for several environments related to real world simulations and competitive video games. However, these techniques require a prior training process that often demands high computational power and can produce unpredictable results. Furthermore, they are trained for specific environments, so it is generally necessary to retrain the model if we want to use it in a different environment or if the environment undergoes significant changes. Moreover, while these techniques are particularly effective at achieving near-optimal behavior, which is beneficial for scoring-focused environments, they are not well-suited for implementing plausible, coherent, or human-like behaviors. \cite{deeprl}

Finally, utility AI involves decision-making models that choose the best decision at each moment based on a utility function or algorithm. When the agent has to make a decision, it calculates the utility of each possible decision and selects the most profitable one. This approach aims to address some problems with behavior trees, providing models that make believable decisions even in unexpected scenarios and are easily extensible and reusable. Similar to the machine learning approach, utility AI also provides a method for achieving high performance, but without having to train the model for each specific environment. However, like machine learning approaches, this technique requires a scoring-focused environment, making it unsuitable for implementing plausible, coherent, or human-like behaviors. \cite{graham2014introduction}

\subsection{Large Language Models}

Language Modeling is the branch of AI that aims to model the human languages. However, since formalizing an entire language seems impossible, this field primarily focuses on predicting (or obtaining a probabilistic set of) words, characters, or groups of words that fit a certain context.

Historically, language modeling has been conducted using $N$-grams, techniques based on statistical distributions that generate the probability of a certain word occurring next to the immediately preceding $N$ words. For instance, a bi-gram provides the probability of a word following its immediately preceding word, and a tri-gram offers the probability of it following the previous two words. However, in the last decade, the field has witnessed a paradigm shift with the emergence of deep learning models that achieve remarkable results in language modeling.

A large language model typically refers to a language model that internally uses deep learning techniques with enormous architectures (in the order of millions of parameters) trained for general purposes. These models aim to generate the next most suitable token (commonly, a character) given a text window.  Their performance and flexibility have rendered them invaluable for numerous natural language processing tasks. Furthermore, by concatenating the generated token to the previous text and re-passing it to the model, we can generate words, sentences, and even entire texts.

The key to the recent acclaim of these models lies in their general-purpose nature, as a single model can generate consistent text across a wide range of tasks and subjects. The release of OpenAI GPT-3 \cite{brown2020language} surprised with its overall performance, becoming a media phenomenon. Since then, a succession of models has continually emerged and improved performance. Some noteworthy examples include Gemini \cite{manyika2023overview}, developed by Google;\footnote{Gemini was initially called Bard, but Google changed its name in 2024} Llama 2 \cite{touvron2023llama} and Llama 3 \cite{llama3}, developed by Meta and whose code was made available for non-commercial use; Vicuna \cite{zheng2023judging}, an open-source model derived from Llama 2; and Mistral \cite{jiang2023mistral}, an independent open-source.

To enhance the accessibility of LLMs, several tools have emerged, simplifying their setup and execution. One widely used tool is the Llama.cpp library \cite{llamacpp}. Furthermore, full-featured software applications for executing LLMs are emerging. An example of this is the desktop application of LM Studio \cite{lmstudio}, which is still in the beta phase but offers many conveniences for downloading, setting up, and running LLMs.

It is also important to mention Hugging Face \cite{huggingface}, the leading machine learning website, which serves as both a social media platform and a repository for machine learning resources. It offers a wide variety of models, datasets, documentation, and tools for utilizing machine learning technologies. Currently, it serves as the primary source of LLMs and related software, housing most existing models, available in various sizes or formats for execution.

In this context, thanks to the considerable capacity of LLMs and the tools developed for their utilization, the use of LLMs in various areas is being explored. Notably, their integration in intelligent agents or NPCs has resulted in interesting use cases. They are being employed for dialogue with users (i.e. the Inworld Origins \cite{origins}, The Matrix Awakens \cite{matrix} and Nvidia Ace \cite{nvidiaAce} technical demos); decision-making based on trust (Suck Up! \cite{suckup}, a video game where the player has to persuade people to enter their home); enhancing the performance of learning agents (Voyager \cite{wang2023voyager}, an intelligent agent playing Minecraft), and even creating complex multi-agent environments with realistic emergent behavior (Generative Agents \cite{generativeagents}).

However, this field is still evolving, and the projects mentioned above primarily serve as demonstrations or small-scale endeavors. While LLMs introduce new techniques in natural language processing, they do not seem to entirely replace classical techniques. Although they offer significant advantages, LLMs also present several problems, such as unpredictability and the risk of generating irrelevant or incoherent responses, commonly referred to as 'hallucinations.' Despite LLMs' responses generally being coherent, it is known that they sometimes generate content that is irrelevant, made-up, or inconsistent with the input data. \cite{huang2023surveyhallucinationlargelanguage} The causes can vary but are usually related to the fact that an LLM is still a machine learning model. This can result from using flawed data during the training process or from overfitting (where the model returns similar outputs when the given prompt closely aligns with the training data).

\subsection{Automated Planning}

AP \cite{ghallab2004automated} is a branch of AI focused on the importance of constructing a trackable plan to solve a problem, rather than merely finding a solution. As these problems often model real-world scenarios, it is essential to create an accurate representation of the environment, known as the world, and a state of the world would encompass the complete definition of a specific state of the environment. An AP problem can be defined by three key elements:

\begin{itemize}
\item The initial state of the world, typically defined as a set of logical predicates.
\item A goal to achieve, usually defined as a set of conditions or logical expressions. The problem is considered solved when a path is found to a state in which all the conditions are satisfied.
\item A set of actions that can be performed to modify the world state. These actions are typically defined with a set of preconditions (the world state required for execution) and effects (the modification of the world state upon execution).
\end{itemize}




\begin{figure}[b]
    \centering
    \includegraphics[width=0.9\linewidth]{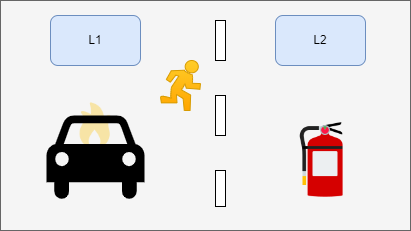}
    \caption{Representation of the initial state of \textit{The Simplified FireFighter Problem}. In location L1 there is a burning car and in location L2 there is an extinguisher. The firefighter is in L1 an its goal is to put out the fire using the extinguisher.}
    \label{fig:monkey-problem}
\end{figure}

Solving an AP problem typically involves two main steps: (1) representing the problem in a formal language interpretable by a program, and (2) obtaining the plan using a solver, an algorithm capable of finding a plan to reach the goal from the initial state. Often, it is necessary to use a program familiar with the formal language and the details of the solver to translate the problem into a model interpretable by the solver.

While there are various approaches to representing planning problems, the current standard planning language is the Planning Domain Definition Language (PDDL) \cite{aeronautiques1998pddl}. PDDL is a general-purpose planning language based on the idea of separating the definition of a problem into two files:

\begin{enumerate}
\item The \textbf{Domain} file, which defines the properties of the world. It specifies the types of objects that exist in the environment, the permissible types of predicates, and all possible actions. Additional components can be found in a PDDL Domain, with the full list available on the \url{planning.wiki} website \cite{planningwiki}.

\item The \textbf{Problem} file, which defines a particular problem within the domain, providing the initial state and goal.
\end{enumerate}

For clarity, we provide an illustrated example, a simplified version of \textit{The FireFighter Problem} (Fig. \ref{fig:monkey-problem}) along with its corresponding PDDL definition (Listings \ref{listing:monkey-domain} and \ref{listing:monkey-problem}).

Finally, note that there is a wide variety of solvers (also known as planners) for solving AP problems, each employing its own strategies and heuristics. Most state-of-the-art solvers include systems for executing PDDL problems, and planning.wiki offers an extensive list of them.\footnote{\url{https://planning.wiki/ref/planners}} Notable planners include GOAP \cite{goap}, ENHSP \cite{scala2016interval}, Metric-FF \cite{hoffmann2003metric}, and Fast Downward \cite{helmert2006fast}, the latter being utilized in this project.

\begin{lstlisting}[caption=Domain Definition file of \textit{The Simplified FireFighter Problem}.  An only action is shown here{,} but the rest of them have an analogous structure., label=listing:monkey-domain]

(define (domain firefighter)
    (:types
        Location Item - object
        Car Extinguisher - Item
    )
    (:predicates
        (In ?obj - Item ?l - Location)
        (FirefighetHas ?obj - Item)
        (IsBurning ?obj - Item)
        (FirefighterIn ?l - Location)
    )
    (:action takeExtinguisher
        :parameters (
            ?e - Extinguisher 
            ?l - Location
        :precondition (and
            (In ?e ?l)
            (FirefighterIn ?l)
        )
        :effect (and
            (not (In ?e ?l))
            (FirefighterHas ?e)
        )
    )
    (:action go_from_to ...)
    (:action putOutFire ...)
)
\end{lstlisting}


\begin{lstlisting}[caption=Problem Definition file of \textit{The Simplified FireFighter Problem}., label=listing:monkey-problem]

(define (problem firefighter-problem)
    (:domain (firefighter))
    (:objects 
        l1 l2 - Location
        e1 - Extingusher
        c1 - Car
    )
    (:init
        (In Car l1)
        (In Extinguisher l2)
        (FirefighterIn l1)
        (IsBurning c1)
    )
    (:goal (not (IsBurning c1))
)
\end{lstlisting}

\section{Agent Architecture}
\label{sec:design}

\begin{figure*}
    \centering
    \includegraphics[width=1\linewidth]{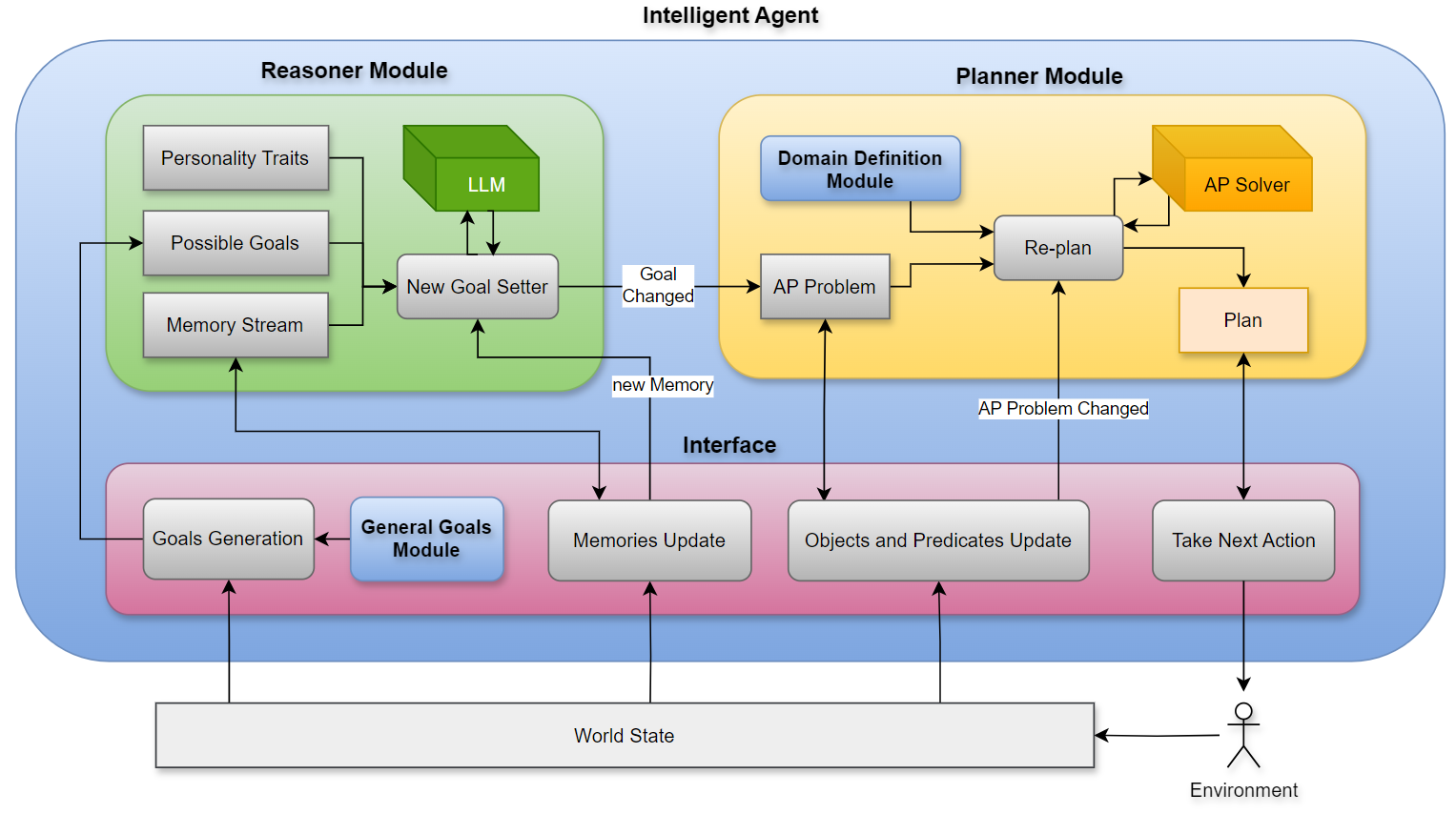}
    \caption{Complete architecture of the Intelligent Agent designed. As we can see, the \textit{Interface} receives the world state's information and uses it to update the possible goals (with the help of the \textit{General Goals} module), the memories of the \textit{Reasoner} and the objects and predicates of the AP Problem. If a new memory has been added, the Interface says the Reasoner to set a new goal, an it is done by calling the LLM. If the AP Problem has changed since the last iteration, the Interface says the Planner to re-plan, which uses the AP Problem (and the AP Domain defined by the \textit{Domain Definition} module) to create a new plan. The interface sequentially take the actions of the generated plan and gives them to the environment.}
    \label{fig:CompleteArchitecture}
\end{figure*}

The complete architecture of the Intelligent Agent is illustrated in Fig. \ref{fig:CompleteArchitecture}, consisting of three main modules: 
\begin{enumerate}
    \item The \textbf{Reasoner}, responsible for managing the memory stream and the LLM calls.
    \item The \textbf{Planner}, responsible for managing the AP problem an the AP Solver calls.
    \item The \textbf{Interface} with the environment, responsible for translating the data-structures, making calls the other modules and returning the actions to the environment.
\end{enumerate}

While the Reasoner and the Planner are context-independent, the Interface module is not, as its task involves direct interaction with the specific characteristics of the environment, requiring dedicated translations and function calls. Additionally, two specific auxiliary modules are also necessary:

\begin{enumerate}
    \item A \textbf{Domain Definition} module. As discussed in Chapter \ref{sec:art}. \textit{State of the Art}, AP problems typically require a prior definition of the domain, which is specific to the environment.
    \item A \textbf{General Goals} module. As outlined in Chapter \ref{sec:sys_desc}. \textit{System Description}, the interface must generate all possible goals from which the Reasoner selects. This module provides information about general goals that an agent could follow (e.g., ``Put out fires'' in The FireFighter Problem), enabling the interface to create specific goals (e.g., ``Put out Fire1'').
\end{enumerate}

In this section, we elaborate on the design considerations of each of these modules.

\subsection{Reasoner}

The Reasoner module stores the relevant information that the LLM needs and implements the corresponding methods to allow the interface to modify it. When the selection of a specific goal is requested to the LLM, this module provides the following information in natural language:

\begin{itemize}
    \item \textbf{Personality traits}: This includes a brief description of the role and expected personality of the specific NPC that is reasoning. It encompasses the name, role (e.g., firefighter or paramedic), and duty of the NPC. This information is provided by the user and should be the sole necessary individual specification to control the particular behavior of each different agent. An example of a personality trait used in The FireFighter Problem is: ``I am 'FireFighter1', a firefighter. My duty is to put out fires and, above all, to save people.''

    \item \textbf{Possible Goals.} This comprises a list of possible specific goals that the NPC can pursue at a given moment. In \textit{The FireFighter Problem}, these goals could include ``Save Peter'' or ''Put out Fire1,`` for example. It must be provided by the interface when a new goal request is made.

    \item \textbf{Memory Stream.} This consists of a list of memories, primarily perceptions that the NPC has experienced, written in natural language. It allows the LLM to understand the particular situation in which it needs to make a decision. The interface frequently updates this module about the world state, and when a new relevant change occurs, it adds it to the memory stream and returns a signal.

\end{itemize}

The main method of this module is the \textbf{New Goal Setter}. It is called by the interface, giving the possible goals list, and utilizes all the previously explained information to generate a prompt to the LLM, asking for the best goal to set. If it is a new goal, the one that the Planner had is modified. If it is the previous one, nothing is done. The decision (whether to keep the goal or change to a new one) is then returned to the interface.

\subsection{Planner}

The Planner module is responsible for generating and maintaining an AP Problem that accurately reflects the state of the world at any given moment. It is also responsible for making the calls to the AP Solver to generate a plan. To fully define the AP Problem, the planner stores the following information:
\begin{itemize}
    \item \textbf{Objects and Predicates.} In AP, a world state is typically defined by a set of objects present in the environment (along with their types) and a set of logical predicates that defines their initial relations or internal states. In our work, we maintain the same approach. 
    
    To keep track of the objects and predicates present in the specific world state at any moment, the interface regularly reports to this module about the world state. When an object or predicate appears or disappears from the environment, the internal AP problem is updated accordingly, and a signal is sent back to the interface.

    \item \textbf{Goal.} This represents the objective that the solver needs to achieve. It is updated by the Reasoner module every time a new goal is established.
\end{itemize}

To generate a new plan, the method \textbf{Re-plan} is utilized. It is called by the interface and utilizes the internal AP Problem definition along with the AP Domain definition (provided by the Domain Definition module) to call the AP Solver, which generates the new plan. This plan is stored as an internal state of the module and replaces the previous one.

\textbf{\textit{Domain Definition Module}}: This auxiliary module provides a specific AP Domain definition of the environment. It is necessary to call the AP Solver and usually cannot be automatically generated using the world state information. Therefore, in such cases, it must be manually defined.

\subsection{Interface}

As we have mentioned, the Interface module is responsible for coordinating the actuation of the other main modules as necessary. Its objective is to generate an intelligent, reactive and plausible behavior using the information provided by the other modules. Although the implementation details may vary based on the specific data structure of the world, the underlying logic remains consistent. It involves an iterative process that repeats the following steps each iteration:

\begin{enumerate}
    \item \textbf{Generate the possible goals list.} Utilizing the General Goals module and the current world state, the Interface module creates a list of specific goals that could be pursued by the agent. 

    Although the General Goals module is explained later in this section, in short, it parses the current world state alongside the possible general goals to provide all the feasible specific goals that could be followed. For instance, if there exists the general goal ``Put out fires'' and there is a fire called ``Fire1'' in the world state, the interface generates the specific goal ``Put out Fire1'' and adds it to the list.

    \item \textbf{Report the new world state to the Reasoner and the Planner}. Both modules process the information and return a signal indicating whether there has been a change in the internal representation of the world.

    \item \textbf{Handle a Re-Think necessity.} If the Reasoner indicates a memories update, the Interface module calls the Reasoner's \textit{New Goal Setter} method. It decides which goal will be set and indicates whether the goal has changed compared to the previous one.
   
    \item  \textbf{Handle a Re-Plan necesity.} If the Planner indicates that the internal world state representation has changed due to new perceptions or if a new goal has been chosen by the Reasoner (i.e., the AP Problem has changed), the Interface module calls the Planner's \textit{Re-plan} method.
    
    \item \textbf{Execute the action}. Finally, the Interface module executes the action at the top of the plan and removes it from the plan.

\end{enumerate}

The rationale behind this procedure is that the Reasoner decides ``What to do'' using the LLM's decision-making capacity, while the Planner decides ``How to do it'' using the AP's planning capacities. This results in a reasonable, intelligent, reactive, and humanized agent with a high degree of autonomy.

\textbf{\textit{General Goals Module}}: This auxiliary module must provide the necessary information to instantiate every general goal. This information includes:

\begin{itemize}
    \item The mechanism to determine if the general goal can be instantiated, on an specific world state, with certain parameters.
    \item The phrase, written in natural language, to provide to the Reasoner to indicate the instantiated goal option.
    \item Similarly to the previous point, the predicate or list of predicates to provide to the Planner if the instantiated goal is chosen.
\end{itemize}

For example, consider a general goal \textit{Heal Persons} with an object as a parameter. In this case, the mechanism to allow instantiation of the goal should ensure that the object is a person and that person is injured. If we attempt to instantiate that goal with the object ``Peter,'' who is currently injured, it should generate a possible goal with the Reasoner's phrase ``Heal Peter'' and the associated Planner's goal predicate ``(healed Peter)''.


\section{Technologies Used}
\label{sec:tech}

To implement the proposed model on a real environment, we have needed, in addition to the environment framework, a tool to set up and run an LLMs and a software to implement the AP Solver. The technologies finally chosen are explained in this section. They are: the Rhymas framework \cite{Sabater-Mir2022} to set up the 3D environment, the LM Studio software \cite{lmstudio} to execute the LLM and the Unified-Planning library \cite{unified_planning_library} to define and solve the AP Problem.

\subsection{Rhymas}
\label{sec:rhymas}

As mentioned in Chapter \ref{sec:use_case}. \textit{Use Case}, Rhymas is a simulation framework developed in collaboration with the ``Escola de Bombers i Protecció Civil de Catalunya.'' It is an under-development framework specifically designed for firefighter training, under which this project is conducted. This framework aims to facilitate large-scale training simulations, using a multi-scale paradigm with autonomous agents and multi-agent system technology to reduce the necessity of additional human intervention during training sessions.

In Rhymas, a simulation comprises a single back-end, which implements the multi-agent-based simulation logic, and several (one or more) front-ends. Each front-end represents a different point of view of the simulation, providing the graphical representation and controlling specific aspects such as perception and physics. Therefore, in Rhymas, every entity is defined by two different objects: one in the back-end, responsible for behavior decisions (acting as the ``brain''), and one in the front-end, which has the sensors and actuators (acting as the ``body''). Consequently, the back-end movements and interactions have a real effect on the front-end simulation, and the front-end changes affect the back-end's internal world state.

The reason behind this architecture is to combine the multi-agent and multi-scale paradigms, so every interaction is conducted through messages. This includes communication between the two parts of an entity, interaction between different entities, and even information exchange between different simulations in a multi-label parallel execution.

In the back-end representation, every entity is seen as an agent, with its own behavior and properties. A world state is composed of all the agents in the scenario, their properties, and the possible relations between them. Dynamic behavior emerges from the interactions between agents, where they can create relations, erase them, or modify some properties of themselves or others. The front-end receives the initial world state from the back-end and initializes a graphical representation of it. To create a realistic simulation of actions and perceptions, we use a state-of-the-art game engine\footnote{In this project's implementation, we are using Unreal Engine 5.0.3}. The engine simulation is responsible for executing the actions indicated by the back-end and informing it about the environment changes.

Under this principle, defining our own environment is as easy as defining the initial world state (scenario) in the back-end. It will be automatically built in the front-end and dynamically change according to the agents' defined behavior. A scenario is defined by the agents initially in the world state, their initial properties, and the initial relations between them.

The Rhymas framework provides a generous list of implemented agent types, which can be modified or used as templates to create new ones. Each type is defined by a Python class, and the most important method for specifying an agent's behavior is the \textbf{run()} method. This method is called for the agents in every iteration, and within it, actions can be taken, including modifying, creating, or erasing relations or properties. 

Thus, to define an agent type, it is only necessary to create its own class and, to integrate it into a scenario, indicate it in the scenario's definition. To illustrate this process, we provide the following simple example of an agent whose behavior consists of walking forward:
\begin{enumerate}
    \item \textbf{Define the agent's behavior.} To do this, we can make a copy of the provided \textit{CommonPerson} implementation, named \textit{ForwardPerson}, and then modify the \textbf{run()} method. (Listing \ref{lst:rhymas-run})

    \item \textbf{Define the Scenario.} Scenarios are defined by a JSON file containing all the necessary information about the initial world state. To add a new agent, navigate to the "agents" dictionary and add a new entry with all the relevant details. As shown in Listing \ref{lst:rhymas-forward_person_scenario}, several key pieces of information need to be specified:
    \begin{itemize}
        \item \textbf{UID}. The unique identifier by which the agent will be recognized during execution.
        \item \textbf{agent\_type}. Defines the behavior the agent will exhibit (set this to ``ForwardPerson'' to assign the desired behavior).
        \item \textbf{running}. Indicates whether the \textit{run()} method is called for that agent (if set to false, the agent will not act autonomously).
        \item \textbf{front\_end\_parameters}. Provides the necessary information to represent the agent in the engine simulation, including at least the position, rotation, scale, and model used.
    \end{itemize}    
\end{enumerate}

Following these steps, defining new agent types and integrating them into Rhymas simulations becomes straightforward. Similarly, relations can be declared within a scenario in an analogous but simpler manner:  unlike agents (which require a Python Class defining their type and behaviour), relations do not, as they do not have an associated behaviour. They simply establish an association between two agents (Listing \ref{lst:rhymas-burning_scenario}). Thereby, enhancing the flexibility and scope of training scenarios.

\begin{lstlisting}[language=python, caption=Run method definition on the ForwardPerson class., label=lst:rhymas-run]

def run(self, tick_counter, sim_time):
    super().run(tick_counter, sim_time)
    self.go_forward()
\end{lstlisting}

\begin{lstlisting}[language=python, caption=Example of a ForwarPerson agent defined on a Scenario JSON., label=lst:rhymas-forward_person_scenario]

"Sim_01_ForwardPerson_0": {
    "UID": "Sim_01_ForwardPerson_0",
    "agent_type": "ForwardPerson",
    "running": "True",
    "front_end_parameters": {
        "position": [0, 0],
        "rotation": [0, 0, 0],
        "map_web_app": {"scale": [1, 1]},
        "unity": {
            "Prefab": "Commonperson"},
        "unreal": {
            "Blueprint": "Commonperson"}
        }
    }
\end{lstlisting}

\begin{lstlisting}[language=python, caption=Example of a Burning relation defined on a Scenario JSON., label=lst:rhymas-burning_scenario]

"Sim_01_Burning_0": {
  "UID": "Sim_01_relation_0",
  "type": "RELBurning",
  "rel_pred": "Sim_01_Fire_0",
  "rel_succ": "Sim_01_CommonCar_0"
},
\end{lstlisting}

\subsection{LM Studio}
\label{sec:lm_studio}

LM Studio is an application for running LLMs. It provides a graphical interface through which a model can be directly downloaded from Hugging Face. Once downloaded, the model can be configured, loaded into memory, and a local server can be run from the interface.

Once this setup is complete, the interface offers an interactive chat tool with the model. The most useful feature, however, is the accessibility using curl or Python, regardless of the specific model being used.  We integrated LM Studio in our software through Python remote calls, whose process is exemplified on Listing \ref{listing:LMStudio-example}. Below, we explain how it works step by step:

\begin{enumerate}
    \item \textbf{Instantiate a client connection to the local server}. The easiest way to do this is using the OpenAI API \cite{openiaApi}. (1)
    \item \textbf{Create the list of messages that the model will receive}. For each message, specify the role of the sender and its content. Typically, this list consists on only two messages: a message from the \textit{system}, providing a general behaviour specification; and a message from the \textit{user}, providing the intended prompt. (2)
    \item \textbf{Make a server request}. Make a request providing the created list of messages and indicating the temperature. As in this case we want solid answers, we use temperature 0 to minimize the randomness of the output. (3)
    \item \textbf{Process the server response.} The server will return a \textit{completion} object with all the relevant information about the call and the response. This includes the model path, the number generated tokens, etc. The most important parameter is the \textit{choices} attribute, which is a list of all generated responses. As multiple responses could be returned from the same call, depending on the model and parameter configuration, they are stored as a list of \textit{choice} items. Each of them contains all the information related to the specific given response, as its index or the finishing generating reason, but the most important one is the \textit{message} attribute, which provides the generated response message. In this case, we always generate just one response, that's why we always take the first element of the array. (4)
\end{enumerate}

\begin{lstlisting}[language=Python, caption=Example of an LM Studio local server call using python., label=listing:LMStudio-example]

(1) from openai import OpenAI

    client = OpenAI(
      base_url="http://localhost:1234/v1", 
      api_key="not-needed")

(2) messages = [
      {"role": "system",  "content": 
       ``You must respect the format''},
      {"role": "user", "content": 
       ``prompt.''} ]

(3) completion = 
    client.chat.completions.create(
    messages=messages, temperature=0)

(4) output = completion.choices[0].message

\end{lstlisting}

\break

Although LM Studio is still in beta, it is already a powerful tool due to the various features it offers: a direct connection with Hugging Face, a local server setup, and an intuitive graphical interface that allows for downloading, configuring, loading, and executing models, all within the same application.

For clarity, we provide an execution example of a simplified version of \textit{The FireFighter Problem}. Using the previous structure, we instantiate the \textit{prompt} field (Listing \ref{listing:LMStudio-firefighter_prompt}). The returned output from the server, as previously said, is a \textit{completion} object. If we extract the actual returned message,  we can read the answer of the model to that prompt (Listing \ref{listing:firefighter_answer-example}).


Finally, note that the local server setup feature is particularly beneficial in our case. In this work, we are already running two simulations (one Rhymas back-end and one Rhymas front-end) and an LLM, which is highly demanding on GPU resources. This feature allows us to distribute the simulation run and model load across different machines, thereby improving execution performance.

\begin{lstlisting}[language=Python, caption=Definition of a prompt that represents a simplified version of \textit{The FireFighter Problem}., label=listing:LMStudio-firefighter_prompt]

prompt = '''
 You are Marc, a Firefighter. Your duty 
 is to put out fires and, above all, 
 to save people.
 You have the following perceptions:
  - There is a fire called f1.
  - There is a person called p1, 
  - p1 is inside of f1.
 What should you do? You must choose 
 only one option:
  1. Save p1
  2. Put out f1
 Indicate the number of the chosen answer.
'''

\end{lstlisting}

\begin{lstlisting}[language=Python, caption=Message given by the LLM in response to the example prompt., label=listing:firefighter_answer-example]

output = completion.choices[0].message 
print(output)

 -> ''' As a firefighter, my priority  
    is to save lives, so I would first 
    assess the situation and determine 
    the best course of action. In this 
    case, there is a person inside the 
    burning building, which means that 
    saving them should be my top priority. 
    Therefore, I would choose option 1:
    Save p1. '''
\end{lstlisting}

\subsection{Unified-Planning}
\label{sec:up}



Despite PDDL being the standard format for writing AP Problems, it presents two main issues: the first is that it requires creating the Domain File and the Problem File in plain text instead of using a manageable data structure, which is a significant handicap when it comes to creating, modifying, or consulting data at runtime; the second is that each AP Solver is included in its own planning system software, which must be externally called and has its own output format, necessitating individual post-processing of each solver's output to adapt it to the desired format.

The AIPlan4EU Project aimed to address these problems by making a European-wide initiative to unify and user-center AP resources. The result of that effort is the Unified-Planning library (UP), a complete Python API that allows for the easy specification of the AP Domain and the AP Problem using consistent Python structures and integrates a pool of AP Solvers that can be directly invoked with these structures, returning the plan in a consistent data structure. Listing \ref{listing:up_example}, shows an example illustrating how to use the library to define and solve \textit{The Simplified FireFighter Problem}, introduced in Chapter \ref{sec:art}. \textit{State of the Art}. Below, we explain how it works step by step:
\begin{enumerate}
    \item[0)] Import the necessary classes and data structures. (1)
    
    \item \textbf{Define the AP Domain}. First, define all the types that the AP Domain will include (2). Then, add the predicates and, in general, every instantiable variable, which are called \textit{fluents} (3). Finally, declare the actions (4). The example shows the definition of the \textit{takeExt} action, which is analogous to the other actions (\textit{goFromTo} and \textit{putOutFire}).

    \item \textbf{Instantiate the Problem Object}. Once all the necessary domain information is defined, instantiate the problem and add the previously declared fluents and actions (the types are implicitly added with them). (5)

    \item \textbf{Define the AP Problem}. Declare all the objects that the AP Problem includes and add them to the instantiated problem (6). Then, define the initial state of the problem by initializing the corresponding instantiated fluents, and declare all the goals that the solver will need to reach (7).

    \item Once the problem instance is completely defined, call a planner to solve it. The parameter specifies which AP solver to use. (8)

\end{enumerate}

The \textit{result} object obtained contains useful information about the execution and always has the same structure regardless of the solver used. It provides information about the status (whether it was solved or not), the solver used, and the generated plan. The resulting plan obtained for the illustrated example is shown in Listing \ref{listing:up_plan}.

The example we have seen is just a simplification of what unified-planning can achieve. This library is very powerful, allowing the solution of more than propositional problems because fluents can be of any type, not only boolean. It is possible to use numerical and temporal fluents or create custom types. The library also has integrated methods to consult and modify instantiated problems and provides tools to implement custom solvers.

\break

\begin{lstlisting}[language=Python, caption=Unified-planning code to solve the \textit{Simplified FireFighter Problem}., label=listing:up_example]

(1)
from unified_planning.shortcuts import *

(2)
Location = UserType('Location')
Item = UserType('Item')
Car = UserType('Car', father=Item)
Ext = UserType('Ext', father=Item)

(3)
In = Fluent(
 'In', BoolType(), obj=Item, l=Location)
IsBurning = Fluent(
 'IsBurning', BoolType(), item=Item)
Has = Fluent(
 'FireFigherHas', BoolType(), obj=Item)
FireFigherIn = Fluent(
 'FireFigherHas', BoolType(), l=Location)

(4)
takeExt = InstantaneousAction(
    'takeExt',  e=Ext, l=Location)
e = takeExt.parameter('e')
l = takeExt.parameter('l')
takeExt.add_precondition(In(e, l))
takeExt.add_precondition(FireFighterIn(l))
takeExt.add_effect(In(e, l), False)
takeExt.add_effect(FireFighterHas(e), True)

(5)
problem = Problem('FireFighter')
problem.add_fluenst(
    [In, IsBurning, FireFighterIn, Has])
problem.add_actions(
    [takeExt, goFromTo, putOutFire])

(6)
l1 = Object('l1', Location)
l2 = Object('l2', Location)
car1 = Object('car1', Car)
ext1 = Object('ext1', Ext)
problem.add_objects([l1, l2, car1, ext1])

(7)
set_initial_value(In(car1, l1), True)
set_initial_value(In(ext1, l2), True)
set_initial_value(FirefighterIn, l1, True)
set_initial_value(IsBurning(car1), True)
problem.add_goal(
    IsBurning(car1).Iff(False)))

(8)
solver = OneshotPlanner('fast-downward')
result = solver.solve(problem)    
\end{lstlisting}

\begin{lstlisting}[language=Python, caption=Plan generated to solve \textit{The Simplified FireFighter Problem}., label=listing:up_plan]

print(result.plan.actions) 
 ->  [ GoFromTo(l1, l2), TakeExt(ext1),
       GoFromTo(l2, l1), PutOutFire(c1) ]
\end{lstlisting}

\section{Implementation}
\label{sec:implementation}

To study the feasibility and performance of the designed intelligent agent and to illustrate its resulting behavior, we implemented it in Rhymas and deployed it in a simple environment, \textit{The FireFighter Problem}. This implementation required addressing five main challenges: integrating the LLM into the system, obtaining AP modeling and solving software, implementing the agent itself, modeling the environment, and finally, setting up and executing the system. In this section, we explain how these issues were addressed and present the final Python implementation of the system. Additionally, to test how its behavior changes with different contexts, we created various scenarios with different numbers of agents and personality traits to observe how they handle \textit{The FireFighter Problem}.

\subsection{LLM Integration}

To make LLM requests, we initially tried some of the LLM-powered web-chat tools provided by various enterprises, such as ChatGPT from OpenAI and Gemini from Google. These options had two main problems: high response times due to web calls and the prompt limitations of the free-use versions. Another alternative was to execute open-source models externally using the HuggingFace API. However, we discovered that only some models could be executed on this API, and the high latency of the online calls remained an issue. Therefore, the more viable option seemed to be downloading and locally executing an open-source model.

In this context, the main tool for LLM management is the Llama.cpp software, but its setup and use are tedious and prone to errors. Along these lines, we ultimately found LM Studio, the software presented in Chapter  \ref{sec:lm_studio}. \textit{LM Studio}. It internally implements Llama.cpp but also offers an easy installation process and a wide range of functionalities.

The main limitation here has been computational power: we are using a mid-range computer with an NVIDIA GeForce RTX 4060 GPU with 8GB of memory, limiting us to smaller versions of the models to maintain reasonable speed. After testing some state-of-the-art models like Llama 2 and Vicuna, we chose Mistral as our final model. Once the model is loaded into memory and the server is running, we can make requests directly from Python, receiving answers in approximately 0.5 seconds.

\subsection{AP Software Selection}

As mentioned in Section \ref{sec:up}. \textit{Unified-Planning}, the PDDL file structure is not convenient for contexts where AP problems are frequently updated. Information should be saved in memory, not written to disk every iteration, and the solver should be executed directly from the program, not through external system calls. In this context, some solvers have alternative Python implementations that allow users to model and execute the problem. We tried a few GOAP implementations but ultimately chose Unified Planning, as it is the best alternative for AP Python implementation, allowing for easy creation and modification of AP problems and providing a wide variety of solvers for execution.

The AP domain of the problem is manually defined as a UP problem Python class and instantiated as a Python object at execution time. When the \textit{Interface} module receives a new perception, it automatically updates the AP problem object's information related to the corresponding world state changes to keep the information up to date. UP also integrates a range of state-of-the-art solvers. We decided to use \textit{Fast Downward}, and its execution is immediate once the problem is defined.

\subsection{Agent Implementation}

To implement the proposed architecture in Python, each module is represented by a class. The implementation described below is based on the \textit{Rhymas} framework and \textit{The FireFighter Problem} scenario, considering that Rhymas defines a world state as a dictionary of agents and peer relations.

\subsubsection{\textbf{Reasoner}} 

This module has a straightforward general implementation of its design. It initializes the memory stream with the personality traits of the specific agent and stores it, providing methods to add and erase memories. When it is necessary to select a new goal, the reasoner makes a request to the \textit{LMStudio} local server executing Mistral, providing the memory stream and asking which of the possible specific goals to follow. If the current goal changes, it is updated in the UP Problem.

\subsubsection{\textbf{Planner}} 

This module also has a straightforward implementation of its design. It provides methods to add and erase objects and relations from the UP Problem instance and to iterate over the plan's actions. When a new plan is needed, the \textit{replan} method calls the UP Planner using the instantiated UP Problem and stores the resulting plan as a list of actions.

\subsubsection{\textbf{General Goals}}

This module is not implemented by a single class but by a set of classes, each representing a general goal. Each General Goal class instantiates all the specific goal objects that can be derived from it. Every goal has an associated phrase that represents it, to be interpreted by the LLM, and a list of AP goals to be completed. For example, in \textit{The FireFighter Problem}, if Peter is in danger, the General Goal \textit{Save} would generate an instantiated object with the phrase \textit{``Take Peter out of the fire''} and the AP goal \textit{Exists(insideOf(``Peter'', ``Safe Zone''))}. This means that if this is the chosen goal, the planner will have to find a plan to move Peter to the safe zone. For this use case, we have implemented a few available general goals, which are: \texttt{ To do nothing}; \texttt{To save someone}; \texttt{To put out a fire}; \texttt{To heal someone}; and \texttt{To call the firefighters}.

\subsubsection{\textbf{Interface}} 

This module focuses on the main \textit{update} method, which is called by the Rhymas agent at every iteration. This method controls the agent's behavior by invoking other module methods and providing them with the corresponding information. It implements the behavior explained in Chapter \ref{sec:design}, \textit{Agent Architecture}:

\begin{lstlisting}[language=python, caption=Iterative method employed to generate all the possible goals, label=lst:goalsGeneration]
possibleGoals = []
For generalGoal in GeneralGoalsModule:
    generalGoal(worldState) -> [goals]
    possibleGoals <- possibleGoals + goals

\end{lstlisting}

\begin{enumerate}
    \item \textbf{Generate possible goals.} The interface iteratively calls all the classes provided by the \textit{General Goals} module, using the given world state as parameter, to generate all possible instances of that class in current given world state (Listing \ref{lst:goalsGeneration}). For example, in an world state where there are two fires (`fire1' and `fire2'), the \textit{`Put Out Fire'} (general goal) class  should generate two possible goals: \textit{``Put Out `fire1'''} and \textit{``Put out `fire2'''}.
    
    As mentioned in the previous section, the \textit{General Goals} module provides the internal mechanisms to generate every possible goal instance, as well as the associated phrase (for the reasoner) and predicates (for the planner) of that goal. 
    
    \item \textbf{Update reasoner memories.} For every agent and relation in the world state, it tries to add it to the reasoner's memories. If it already exists, nothing happens; if it is new, it is added to the Memory Stream. Additionally, every memory of the reasoner is checked to see if it still exists. If it has disappeared, it is also indicated to the memory stream. For example, when the agent perceives the fire for the first time, the memory \textit{"There is a fire called 'Sim\_01\_Fire\_0'"} is added to the Memory Stream. Furthermore, if the fire is extinguished, the memory \textit{"There is a fire called ‘Sim\_01\_Fire\_0’: is no longer true”} is added.

    \textit{In short}: it compares the world state with the reasoner's memories and updates them with the changes.

    \item \textbf{Update planner information.} A process similar to the previous one is followed: for every agent, it attempts to add it as an object in the UP Problem; for every relation, it tries to add it as a predicate. If it already exists, nothing happens; if it is new, it is added. Conversely, through an analogous process, it checks if any objects or relations in the UP Problem have disappeared from the world state, in which case they are removed.

    \textit{In short}: it compares the world state with the UP Problem and updates it with the changes.

    \item \textbf{Set a new goal}. If there is any new memory, it calls the reasoner's new goal setter method.

    \item \textbf{Re-plan}. If the goal or planner information has changed, It calls the planner's replan method.

\end{enumerate}

To integrate this intelligent agent architecture into Rhymas, we need to implement it in an agent class. As the execution flow is derived from the Interface's \textit{update} method, the agent class only needs to call it every iteration with the new world state. To do this, we primarily need to redefine the \textit{run()} method of its class, which will be executed at every iteration. We have selected the existing \textit{AGFireFighter} Rhymas agent and modified its \textbf{run()} method with the following behavior:

\begin{enumerate}
    \item In the first iteration, the interface is initialized (the interface is in charge of initializing the rest of the modules).

    \item In subsequent iterations, it calls the update method of the interface, which returns the action to execute.

    \item If the action is not \textit{None}, the environment executes it.
\end{enumerate}

\begin{figure}[t]
    \centering
    \includegraphics[width=0.9\linewidth]{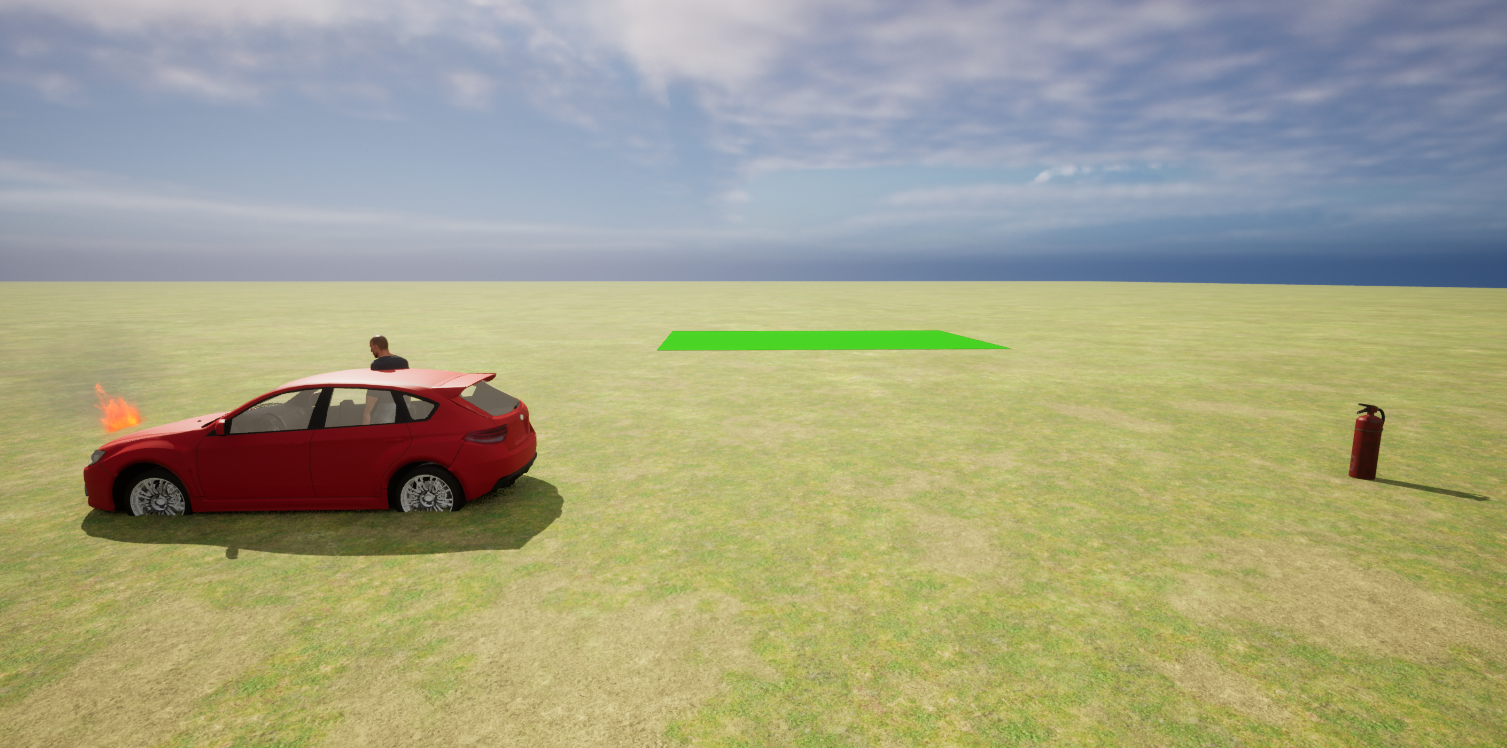}
    \caption{Visualization of \textit{The FireFighter Problem} on Rhymas, using the Unreal simulation. As can be seen, on the left there is the burning car with a person inside, on the right there is the fire extinguisher and in the background there is the safe zone, represented as a green square.}
    \label{fig:unreal-scenario}
\end{figure}

We use the predefined \textit{AGFireFighter} class because it already implements all the necessary methods to give our agent the capability of interacting with the environment as needed the solve \textit{The FireFighter Problem}. These methods include functions such as `go\_to(object\_uid)', `grab(object\_uid)', `em\_extinguish\_fire(fire\_uid)' and `drop(object\_uid)', whose purposes are self-explanatory. The class also provides several attributes that store useful internal state information, such as `em\_current\_action' (which indicates the action that the agent is currently performing) or `em\_fire\_uid' (which identifies the fire the agent is extinguishing, if applicable). This information can be very useful for simplifying the control flow of actions.

\subsection{Environment Modelling}

In order to illustrate the behavior of the intelligent agent, we have also modeled \textit{The FireFighter Problem} in Rhymas. Defining a Rhymas scenario requires to define all the initial agents and the relations. In this case, as depicted in Fig. \ref{fig:firefighetScheme} (Chapter \ref{sec:introduction}. \textit{Introduction}), we have five initial agents and two relations: one between the car and the person, who is \textit{inside of} it; and one between the car and the fire, which is \textit{burning} it. In Rhymas, all the necessary information about the agents ant the relations must be provided by a JSON file. During execution, the JSON definition of this problem is automatically translated into its respective Unreal simulation at execution time (see Fig. \ref{fig:unreal-scenario}), without the necessity of any additional declarations. Since Rhymas' syntax specifications are not particularly relevant to our work, and all agents and relations are instantiated similarly, we provide an example of each one, solely for illustration purposes (see Listings \ref{lst:rhymas-agent_example} and \ref{lst:rhymas-relation_example}). 

\break

\begin{lstlisting}[language=python, caption=Example of an agent instantiation in Rhymas{,} the fire extinguisher in particular. As we can see{,} some specific parameters as \textit{can\_be\_moved} can be especified on the initialization., label=lst:rhymas-agent_example]

"Sim_01_FireExtinguisher_0": {
    "UID": "Sim_01_FireExtinguisher_0",
    "agent_type": "AGFireExtinguisher",
    "running": "True",
    "front_end_parameters": {
        "position": [-10.0, 20.32, -0.55],
        "rotation": [0, 90, 0],
        "map_web_app": {
          "scale": [1.3, 3.6],
          "can_be_moved": true
          }
        "unreal": {
          "Blueprint": "FireExtinguisher"}
        }
    }
\end{lstlisting}

\begin{lstlisting}[language=python, caption=Example of a relation instantiation in Rhytmas{,} the burning relation between the fire and the car in particular. As we can see{,} the initialization is very simple{,} being only needed the related agents and the relation type, label=lst:rhymas-relation_example]

"Sim_01_Burning_0": {
  "UID": "Sim_01_relation_0",
  "type": "RELBurning",
  "rel_pred": "Sim_01_Fire_0",
  "rel_succ": "Sim_01_CommonCar_0"
}

\end{lstlisting}


\subsection{System Setting Up}

To execute the implemented system, it is necessary to simultaneously run the LM Studio local server with the Mistral model and the Unreal Engine simulation of the environment. Both of these components require a high percentage of GPU potential and memory, making it impractical to run them on a single mid-range computer, taking several tens of seconds or even minutes to complete each iteration of the simulation. Fortunately, as explained in Chapter \ref{sec:rhymas}. \textit{Rhymas}, the Rhymas software is divided into two modules: the Python back-end, responsible for multi-agent behaviour and interactions, and the Unreal Engine front-end, controlling the physics and graphic simulation. These modules communicate with each other via websockets, allowing them to be executed on different machines. Therefore, we opted to run the front-end simulation on another similar computer, alleviating the computational burden on the main machine and resulting in significantly improved speed performance.

In summary, the system setup involves one machine executing the Rhymas back-end and the Mistral model, connected via web requests to another machine running the Unreal Engine with the Rhymas front-end simulation.

\subsection{Testing Scenarios}

We have created several different scenarios in which we introduce various intelligent agents into \textit{The FireFighter Problem}, each utilizing our architecture. The only variation among these agents is the prompt used to define their ``personality traits,'' aimed at observing the different behaviors that emerge (see Appendix \ref{appendix}). 

We employed five different configurations for the agents: the person inside the car (CI), an external common person (CO), a firefighter prioritizing saving people (FP), a firefighter prioritizing putting out fires (FF), and a paramedic (PA). We evaluated the behavior of these agents in different situations, beginning with a single agent and expanding up to four simultaneously, to observe their performance in a multi-agent context. All the testing scenarios are listed in Table \ref{t3}.

\begin{table}[h]
\caption{Summary of all the scenarios used to test the behaviour of the developed architecture.}
\label{t3}

\centering
\begin{tabular}{rccccc}
\hline
\multicolumn{1}{l}{} & \multicolumn{5}{c}{\textbf{Scenarios}}              \\ \hline 
\textbf{One Agent}   & CI                & CO        & FP        & FF & PA \\
\textbf{Two Agents}  & FP \& FF         & FP \& PA & CI \& FP &    &    \\
\textbf{Four Agents} & CI, FP, FF \& PA &           &           &    &    \\ \hline
\\
\end{tabular}
\end{table}

\section{Discussions and Future Work}
\label{sec:discussion}

\subsection{Discussions}

After conducting the experiments, we have confirmed that the developed intelligent agent exhibits plausible and coherent behavior: it consistently attempts to rescue the person trapped inside the car or provide as much assistance as possible (for example, in the scenario of a solo common person, if there are no firefighters, it will call for them). We consider this outcome a significant success, demonstrating the agent's ability to autonomously select and pursue goals based on the specific context without requiring explicit decision-making instructions.

Ideally, users should be able to control the reasoning logic and priorities of the NPC by specifying individual characteristics in its personality traits. This mechanism generally works well, observing changes in the NPC's decision tendencies based on their personality. However, full control over the agent's decisions remains elusive, so these traits can only generate some decision propensity. This can be evidenced in the disparity between the two firefighters: the one prioritizing saving people \textbf{always} does so, while the one prioritizing putting out fires does so \textbf{most} of the time but occasionally decides to save the person instead. Similarly, even the common person and the paramedic have ended up prioritizing saving the person, despite their personality traits indicating otherwise. For example, in the case of the common person, to achieve the desired behaviour it was necessary to assign a personality trait as detailed and strict as the following: ``\textit{My duty is to prioritize safety above all. I want to help, but if there is someone more qualified to do it, I won't do anything that could endanger myself}.''

Nevertheless, we have observed that the agent consistently acts in accordance with its designated personality traits. For example, considering the firefighter whose primary duty is to save people in a context where there is only fire, with no one in immediate danger, the agent will prioritize extinguishing the fire, as expected. Thus, we may conclude that the agent acts based on personality traits and exhibits some degree of ``common sense.'' This inherent ``common sense'' explains why the agent is not always faithful to its personality traits, which makes sense given that the LLM's decision-making ability is grounded in its training data, which should not be assumed to be less crucial than the prompt provided for reasoning.

This feature is neither inherently positive nor negative; rather, it is an important characteristic to consider when utilizing an LLM for reasoning or decision-making tasks. On one hand, it provides consistent, plausible, and coherent behaviour; however, on the other hand, its behavior may not easily align with highly personalized desires or instructions.

\subsection{Future Work}

As we have observed, utilizing an LLM for decision-making offers several positive features and considerable potential. However, we have also encountered some drawbacks, specially problematic when implementing the entire reasoning module for intelligent agents, as the uncertainty and inflexible behavior usually are undesirable traits in NPCs of simulations or video games. Therefore, it would be valuable to explore hybridizing this approach with other classical techniques. For instance, an intriguing extension of this work could involve studying the behavior of a decision tree, where some of the leaves are calls to the LLM.

The results obtained in this study are influenced by the use of the Mistral model and its behavior, as well as limited by computational power. In future work, it would be interesting to explore the potential of different models, especially larger ones, to address this problem. Additionally, investigating the behavior of this architecture in larger and more complex environments would be highly valuable.

Furthermore, the developed architecture could be extended in various directions, such as expanding the planner to include temporal considerations or introducing an automatic generator for the domain module.

Finally, a significant avenue for future work would be the development of a more generalized interface module, in order to transform the system into a framework for developing hybrid modular intelligent agents across different types of simulated environments.

\section{Conclusion}
\label{sec:conclusion}

In this study, we have presented the design and implementation of a new intelligent agent proposal, leveraging the decision-making capabilities of LLMs to select the desired goals and the abilities of AP to dynamically generate plans to achieve them. Our preliminary tests demonstrate promising results: despite utilizing a constrained LLM due to computational constraints, the NPCs consistently exhibit the expected plausible human-like behavior, suggesting a promising direction for further investigation. Future research endeavors could involve evaluating the agent's performance with more powerful LLMs, which may help address the fidelity issues observed. Moreover, exploring the combination of this approach with other traditional methods could lead to the development of a reasoning model that combines the strengths of both technologies.

\bibliographystyle{vancouver}
\bibliography{references}

\appendix

\subsection{Prompts generated on the Testing Scenarios}
\label{appendix}

\begin{figure}[h]
    \centering
    \fcolorbox{black}{white}{    \includegraphics[width=0.95\linewidth]{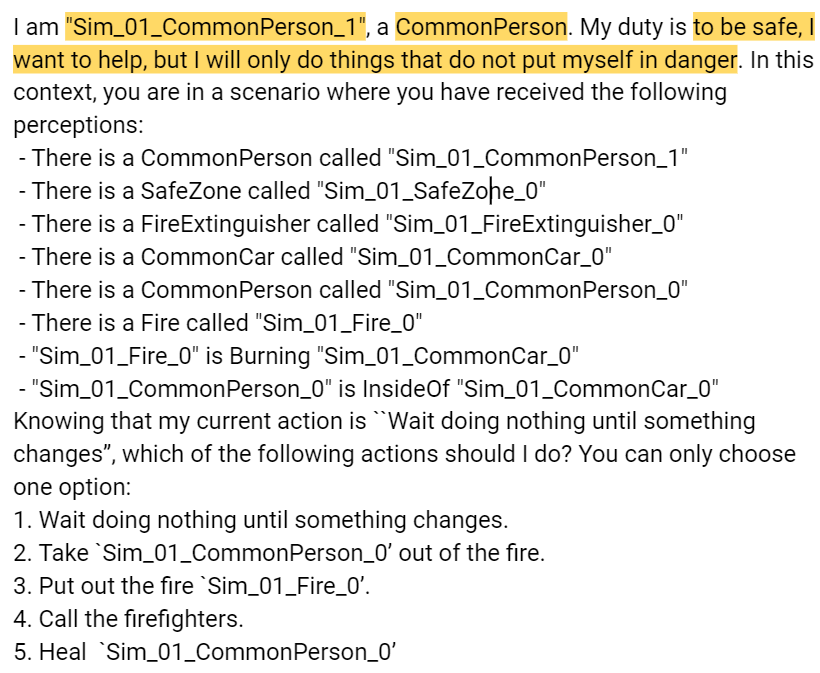}}

    \caption{Example of a prompt generated at the initial state of \textit{The FireFighter Problem}. As the problem is always the same one, the initial state is similar on every scenario.}
    \label{fig:promptExample}
\end{figure}

In Fig. \ref{fig:promptExample}, an example of how the generated prompts look is shown. They are always composed of the following elements: the name and type of the agent, its personality traits, its Memory Stream, the current action that it is executing, and all the possible specific goals to follow. There are highlighted the unique personal characteristics of the agent: its name, type, and personality traits. For each instantiated agent, the user only needs to specify its personality traits to make it act in the desired manner.

\end{document}